%% file: main.tex
\documentclass{article}

\usepackage{microtype}
\usepackage{graphicx}
\usepackage{subcaption}
\usepackage{booktabs} 

\usepackage{hyperref}

\usepackage[accepted]{icml2019}

\icmltitlerunning{CAVIA: Fast Context Adaptation via Meta-Learning}

\usepackage{amsmath, amssymb}
\usepackage{graphicx}
\graphicspath{ {images/} }
\usepackage{multirow}
\usepackage{listings}
\usepackage{spverbatim}
\usepackage{graphbox}  

\begin{document}

\twocolumn[
\icmltitle{Fast Context Adaptation via Meta-Learning}



\icmlsetsymbol{equal}{*}

\begin{icmlauthorlist}
	\icmlauthor{Luisa Zintgraf}{ox}
	\icmlauthor{Kyriacos Shiarlis}{ox,ll}
	\icmlauthor{Vitaly Kurin}{ox,ll}
	\icmlauthor{Katja Hofmann}{msr}
	\icmlauthor{Shimon Whiteson}{ox,ll}
\end{icmlauthorlist}

\icmlaffiliation{ox}{University of Oxford}
\icmlaffiliation{ll}{Latent Logic}
\icmlaffiliation{msr}{Microsoft Research}

\icmlcorrespondingauthor{Luisa Zintgraf}{luisa.zintgraf@cs.ox.ac.uk}

\icmlkeywords{Machine Learning, ICML}

\vskip 0.3in
]



\printAffiliationsAndNotice{}  

\begin{abstract}
	We propose CAVIA for meta-learning, a simple extension to MAML that is less prone to meta-overfitting, easier to parallelise, and more interpretable.
	CAVIA partitions the model parameters into two parts: 
	{\it context parameters} that serve as additional input to the model and are adapted on individual tasks, and {\it shared parameters} that are meta-trained and shared across tasks.
	At test time, only the context parameters are updated, leading to a low-dimensional task representation.
	We show empirically that CAVIA outperforms MAML for regression, classification, and reinforcement learning.
	Our experiments also highlight  weaknesses in current benchmarks, in that the amount of adaptation needed in some cases is small.
\end{abstract}
%

\input{core/introduction.tex}

\input{core/background.tex}

\input{core/contribution.tex}

\input{core/relatedwork.tex}
\input{core/experiments.tex}

\input{core/discussion.tex}

\section*{Acknowledgements}
We thank Frans Oliehoek, Wendelin Boehmer, Mark Finean and Joost van Amersfoort for useful discussions and feedback.
We would also like to thank Chelsea Finn, Jackie Loong and Tristan Deleu for their open-sourced MAML implementations.
The NVIDIA DGX-1 used for this research was donated by the NVIDIA corporation.
L. Zintgraf is supported by the Microsoft Research PhD Scholarship Program.
Vitaly Kurin is supported by the Samsung R\&D Institute UK studentship.
This project has received funding from the European Research Council (ERC) under the European Union's Horizon 2020 research and innovation programme (grant agreement number 637713).
We thank the anonymous reviewers for their valuable feedback which helped us improve this paper.

\bibliography{main}
\bibliographystyle{icml2019}

\appendix

\newpage
\input{core/appendix.tex}

\end{document}

%% file: core/introduction.tex
\section{Introduction}

The challenge of fast adaptation in machine learning is to learn on previously unseen tasks {\it fast} and with {\it little data}.
In principle, this can be achieved by leveraging knowledge obtained in other, related tasks. 
However, the best way to do so remains an open question.
We are interested in the meta-learning approach to fast adaptation, i.e., {\it learning how to learn} on unseen problems/datasets within few shots.

One approach for fast adaptation is to use gradient-based methods: at test time, only one or a few gradient update steps are performed to solve the new task, using a task-specific loss function.
{\it Model agnostic meta learning} (MAML) \citep{finn2017model} is a general and powerful gradient-based meta-learning algorithm, which learns a model initialisation that allows fast adaptation at test time.
Given that MAML is model-agnostic, it can be used with any gradient-based learning algorithm, and a variety of methods build on it (a.o., \citet{lee2018gradient, li2017meta, al2017continuous, grant2018recasting}).

MAML is trained with an interleaved training procedure, comprised of inner loop and outer loop updates that operate on a batch of related tasks at each iteration.
In the inner loop, MAML learns task-specific network parameters by performing one gradient step on a task-specific loss.
Then, in the outer loop, the model parameters from {\it before} the inner loop update are updated to reduce the loss {\it after} the inner loop update on the individual tasks.
Hence, MAML learns a model initialisation that can generalise to a new task after only a few gradient updates at test time.
One drawback of MAML is meta-overfitting: since the entire network is updated on just a few data points at test time, it can easily overfit \citep{mishra2018simple}. 

In this paper, we propose an alternative to MAML which is more interpretable and less prone to overfitting, without compromising performance.
Our method, {\it fast \textbf{c}ontext \textbf{a}daptation \textbf{via} meta-learning} (CAVIA) learns a single model that adapts to a new task via gradient descent by updating only a set of input parameters at test time, instead of the entire network.
These inputs, which we call {\it context parameters} $\phi$ (see Figure \ref{fig:overview}), can be interpreted as a task embedding that modulates the behaviour of the model. 
We confirm empirically that the learned context parameters indeed match the latent task structure.
Like MAML, our method is model-agnostic, i.e., it can be applied to any model that is trained via gradient descent. 

CAVIA is trained with an interleaved training procedure similar to MAML: in the inner loop only the context parameters $\phi$ are updated, and in the outer loop the rest of the model parameters, $\theta$, are updated (which requires backpropagating through the inner-loop update). 
This allows CAVIA to explicitly optimise the task-independent parameters $\theta$ for good performance across tasks, while ensuring that the task-specific parameters $\phi$ can quickly adapt to new tasks at test time.

The separation of parameters into task-specific and task-independent parts has several advantages.
First, the size of both components can be chosen appropriately for the task.
The network parameters $\theta$ can be made expressive enough without overfitting to a single task in the inner loop, which MAML is prone to.
Furthermore, for many practical problems we have prior knowledge of which aspects vary across tasks and hence how much capacity $\phi$ should have.

Second, CAVIA is significantly easier to parallelise compared to MAML: learning task-specific context parameters for a batch of tasks can be parallelised in the inner loop.
Other benefits are that parameter copies are not necessary which saves memory writes; we do not need to manually access and perform operations on the network weights and biases to set up the computation graphs; and CAVIA can help distributed machine learning systems, where the same model is deployed to different machines and we wish to learn different contexts concurrently. 

CAVIA is conceptually related to embedding-based approaches for fast adaptation such as {\it conditional neural processes} (CNPs) \citep{garnelo2018conditional} and \textit{meta-learning with latent embedding optimisation} (LEO) \cite{rusu2018meta}. 
These share the benefit of learning a low-dimensional representation of the task, which has the potential to lead to greater interpretability compared to MAML.
In contrast to existing methods, CAVIA uses the same network to learn the embedding (during a backward pass) and make predictions (during a forward pass). 
Therefore CAVIA has fewer parameters to train, but must compute higher-order gradients during training.

Our experiments show that CAVIA outperforms MAML and CNPs on regression problems, can outperform MAML on a challenging classification benchmark by scaling up the network without overfitting, and outperforms MAML in a reinforcement learning setting while adapting significantly fewer parameters.
We also show that CAVIA is robust to hyperparameters and demonstrate that the context parameters represent meaningful embeddings of tasks.
Our experiments also highlight a weakness in current benchmarks in meta-learning, in that the amount of adaptation needed is small in some cases, confirming that task inference and multitask learning are enough to do well.

%% file: core/background.tex
\section{Background}

Our goal is to learn models that can quickly adapt to new tasks with little data. 
Hence, learning on the new task is preceded by meta-learning on a set of related tasks. 

\subsection{Problem Setting}

In few-shot learning problems, we are given distributions over training tasks $p_\text{train}(\mathcal{T})$ and test tasks $p_\text{test}(\mathcal{T})$.
Training tasks can be used to learn how to adapt fast to any of the tasks with little per-task data, and evaluation is then done on (previously unseen) test tasks.
Unless stated otherwise, we assume that $p_\text{train}=p_\text{test}$ and refer to both as $p$.
Tasks in $p$ typically share some structure, so that transferring knowledge between tasks can speed up learning. 
During each meta-training iteration, a batch of $N$ tasks $\mathbf{T} = \{\mathcal{T}_i\}_{i=1}^N$ is sampled from $p$.

\textbf{Supervised Learning.}
Supervised learning learns a model $f:x\mapsto \hat{y}$ that maps data points $x\in\mathcal{X}$ that have a true label $y\in\mathcal{Y}$ to predictions $\hat{y}\in\mathcal{Y}$. 
A task
	$\mathcal{T}_i = (\mathcal{X}, \mathcal{Y}, \mathcal{L}, q)$ is a tuple
where $\mathcal{X}$ is the input space, $\mathcal{Y}$ is the output space, $\mathcal{L}(y, \hat{y})$ is a task-specific loss function, and $q(x, y)$ is a distribution over labelled data points.
We assume that all data points are drawn i.i.d.\ from $q$.
Different tasks can be created by changing any element of $\mathcal{T}_i$.

Training in supervised meta-learning proceeds over meta-training iterations, where for each $\mathcal{T}_i\in\mathbf{T}$, we sample two datasets $\mathcal{D}_i^\text{train}$ and $\mathcal{D}_i^\text{test}$ from $q_{\mathcal{T}_i}$:
\begin{equation}
\mathcal{D}_i^\text{train} = \{ (x, y)^{i, m}\}_{m=1}^{M_i^\text{train}},~~~
\mathcal{D}_i^\text{test} = \{ (x, y)^{i, m}\}_{m=1}^{M_i^\text{test}},
\end{equation}
where $(x, y)\sim q_{\mathcal{T}_i}$ and $M_i^\text{train}$ and $M_i^\text{test}$ are the number of training and test datapoints.
The training data is used to update $f$, and the test data is then used to evaluate how good this update was, and adjust $f$ or the update rule accordingly.

\textbf{Reinforcement Learning.} Reinforcement learning (RL) learns a policy $\pi$ that maps states $s\in\mathcal{S}$ to actions $a\in\mathcal{A}$. 
Each task corresponds to a {\it Markov decision process} (MDP): a tuple
	$\mathcal{T}_i = (\mathcal{S},  \mathcal{A}, r,  q, q_0)$,
where $\mathcal{S}$ is a set of states, $\mathcal{A}$ is a set of actions, $r(s_t, a_t, s_{t+1})$ is a reward function, $q(s_{t+1} | s_t, a_t)$ is a transition function, and $q_0(s_0)$ is an initial state distribution. The goal is to maximise the expected cumulative reward $\mathcal{J}$ under $\pi$,
\begin{equation} \label{eq:rl_objective}
	\mathcal{J}(\pi) = \mathbb{E}_{q_0, q, \pi} \left[ \sum_{t=0}^{H-1} \gamma^t r(s_t, a_t, s_{t+1}) \right],
\end{equation}
where $H\in\mathbb{N}$ is the horizon and $\gamma \in [0,1]$ is the discount factor.
During each meta-training iteration, for each $\mathcal{T}_i\in\mathbf{T}$, we first collect a trajectory 
\begin{align*}
\tau_i^\text{train} = \{ 
& s_0, a_0, r_0, s_1, a_1, r_1, \dotsc, \\
& s_{M_i^\text{train}-1}, a_{M_i^\text{train}-1}, r_{M_i^\text{train}-1}, s_{M_i^\text{train}} \},
\end{align*}
where the initial state $s_0$ is sampled from $q_0$, the actions are chosen by the current policy $\pi$, the state transitions according to $q$, and $M_i^\text{train}$ is the number of environment interactions.
We unify several episodes in this formulation: if the horizon $H$ is reached within the trajectory, the environment is reset using $q_0$.
Once the trajectory is collected, this data is used to update the policy.
Another trajectory $\tau^\text{test}_i$ is then collected by rolling out the updated policy for $M_i^\text{test}$ time steps. 
This test trajectory is used to evaluate the quality of the update on that task, and to adjust $\pi$ or the update rule accordingly.

Evaluation for both supervised and reinforcement learning problems is done on a new (unseen) set of tasks drawn from $p$. 
For each such task, the model is updated using $\mathcal{L}$ or $\mathcal{J}$ and only a few data points ($\mathcal{D}^\text{train}$ or $\tau^\text{train}$). 
Performance of the updated model is reported on $\mathcal{D}^\text{test}$ or $\tau^\text{test}$.

\subsection{Model-Agnostic Meta-Learning}

One method for few-shot learning is {\it model-agnostic meta-learning} \citep[MAML]{finn2017model}. 
MAML learns an initialisation for the parameters $\theta$ of a model $f_\theta$ such that, given a new task, a good model for that task can be learned with only a small number of gradient steps and data points.
In the inner loop, MAML computes new task-specific parameters $\theta_i$ (starting from $\theta$) via one gradient update\footnote{\label{footnote:gradsteps}We outline is MAML for one gradient update step and the supervised learning setting, but it can be used with several gradient update steps and for reinforcement learning  problems as well.},
\begin{equation} \label{eq:maml_inner}
	\theta_i = 
	\theta - 
	\alpha \nabla_\theta 
	\frac{1}{M^i_\text{train}} 
	\sum\limits_{(x, y) \in \mathcal{D}^\text{train}_i}
	\mathcal{L}_{\mathcal{T}_i}(f_\theta(x), y) \ .
\end{equation}
For the meta-update  in the outer loop, the {\it original} model parameters $\theta$ are then updated with respect to the performance after the inner-loop update, i.e.,
\begin{equation} \label{eq:maml_outer}
	\theta \leftarrow 
	\theta - 
	\beta \nabla_\theta 
	\frac{1}{N} 
	\sum\limits_{\mathcal{T}_i \in \mathbf{T}}
	\frac{1}{M^i_\text{test}}
	\sum\limits_{(x, y) \in \mathcal{D}^\text{test}_i}
	\mathcal{L}_{\mathcal{T}_i}(f_{\theta_i}(x), y) \ .
\end{equation}
The result of training is a model initialisation $\theta$ that can be adapted with just a few gradient steps to any new task drawn from $p$. 
Since the gradient is taken with respect to the parameters $\theta$ before the inner-loop update (\ref{eq:maml_inner}), the outer-loop update (\ref{eq:maml_outer}) involves higher order derivatives of $\theta$.

%% file: core/contribution.tex
\section{CAVIA}
\label{sec:contribution}

We propose {\it fast \textbf{c}ontext \textbf{a}daptation \textbf{via} meta-learning} (CAVIA), which partitions the model parameters into two parts: context parameters $\phi$  are adapted in the inner loop for each task, and parameters $\theta$ are meta-learned in the outer loop and shared across tasks.

\subsection{Supervised Learning}

At every meta-training iteration and for the current batch $\mathbf{T}$ of tasks, we use the training data $\mathcal{D}^\text{train}_i$ of each task $\mathcal{T}_i\in\mathbf{T}$ as follows.
Starting from a fixed value $\phi_0$ (we typically choose $\phi_0=\mathbf{0}$; see Section \ref{sec:method_initialisation}), we learn task-specific parameters $\phi_i$ via one gradient update:
\begin{equation} \label{eq:inner_update}
	\phi_i = 
	\phi_0 - 
	\alpha 
	\nabla_{\phi}
	\frac{1}{M_i^\text{train}}
	\sum\limits_{(x, y) \in \mathcal{D}_i^\text{train}} 
	\mathcal{L}_{\mathcal{T}_i} (f_{\phi_0, \theta}(x), y).
\end{equation}
While we only take the gradient with respect to $\phi$, the updated parameter $\phi_i$ is also a function of $\theta$, since during backpropagation, gradients flow through the model. 
Given updated parameters $\phi_i$ for all sampled tasks, we proceed to the meta-learning step, in which 
$\theta$ is updated:
\begin{equation}
	\theta \leftarrow \theta - 
	\beta
	\nabla_\theta 
	\frac{1}{N}
	\sum\limits_{\mathcal{T}_i \in \mathbf{T}}
	\frac{1}{M_i^\text{test}}
	\sum\limits_{(x, y) \in \mathcal{D}_i^\text{test}}
	\mathcal{L}_{\mathcal{T}_i} (f_{\phi_i, \theta}(x), y) \ .
\end{equation}
This update includes higher order gradients in $\theta$ due to the dependency on (\ref{eq:inner_update}).
At test time, \textit{only} the context parameters are updated using Equation  (\ref{eq:inner_update}), and $\theta$ is held fixed.

\subsection{Reinforcement Learning}

During each iteration, for a current batch of MDPs $\mathbf{T}=\{ \mathcal{T}_i \}_{i=1}^N$, we proceed as follows. 
Given $\phi_0$,
we collect a rollout $\tau_i^\text{train}$ by executing the policy $\pi_{\phi_0, \theta}$.
We then compute task-specific parameters $\phi_i$ via one gradient update:
\begin{equation} \label{eq:inner_update_rl}
\phi_i = 
\phi_0 + 
\alpha
\nabla_\phi 
\tilde{\mathcal{J}}_{\mathcal{T}_i} (\tau_i^\text{train}, \pi_{\phi_0, \theta}),
\end{equation}
where $\tilde{\mathcal{J}}(\tau, \pi)$ is the objective function of any gradient-based reinforcement learning method that uses trajectories $\tau$ produced by a parameterised policy $\pi$ to update that policy's parameters.
After updating the policy, we collect another trajectory $\tau_i^\text{test}$ to evaluate the updated policy, where actions are chosen according to the updated policy $\pi_{\phi_i, \theta}$. 

After doing this for all tasks in $\mathbf{T}$, the meta-update step updates $\theta$ to maximise the average performance across tasks (after individually updating $\phi$ for them),
\begin{equation}
\theta \leftarrow \theta + 
\beta
\nabla_\theta 
\frac{1}{N}
\sum\limits_{\text{MDP}_i \in \mathbf{T}}
\tilde{\mathcal{J}}_{\mathcal{T}_i} (\tau_i^\text{test}, \pi_{\phi_i, \theta}).
\end{equation}
This update includes higher order gradients in $\theta$ due to the dependency on (\ref{eq:inner_update_rl}).

\subsection{Conditioning on Context Parameters} \label{sec:method_conditioning}

\begin{figure}
	\centering
	\includegraphics[width=0.6\columnwidth]{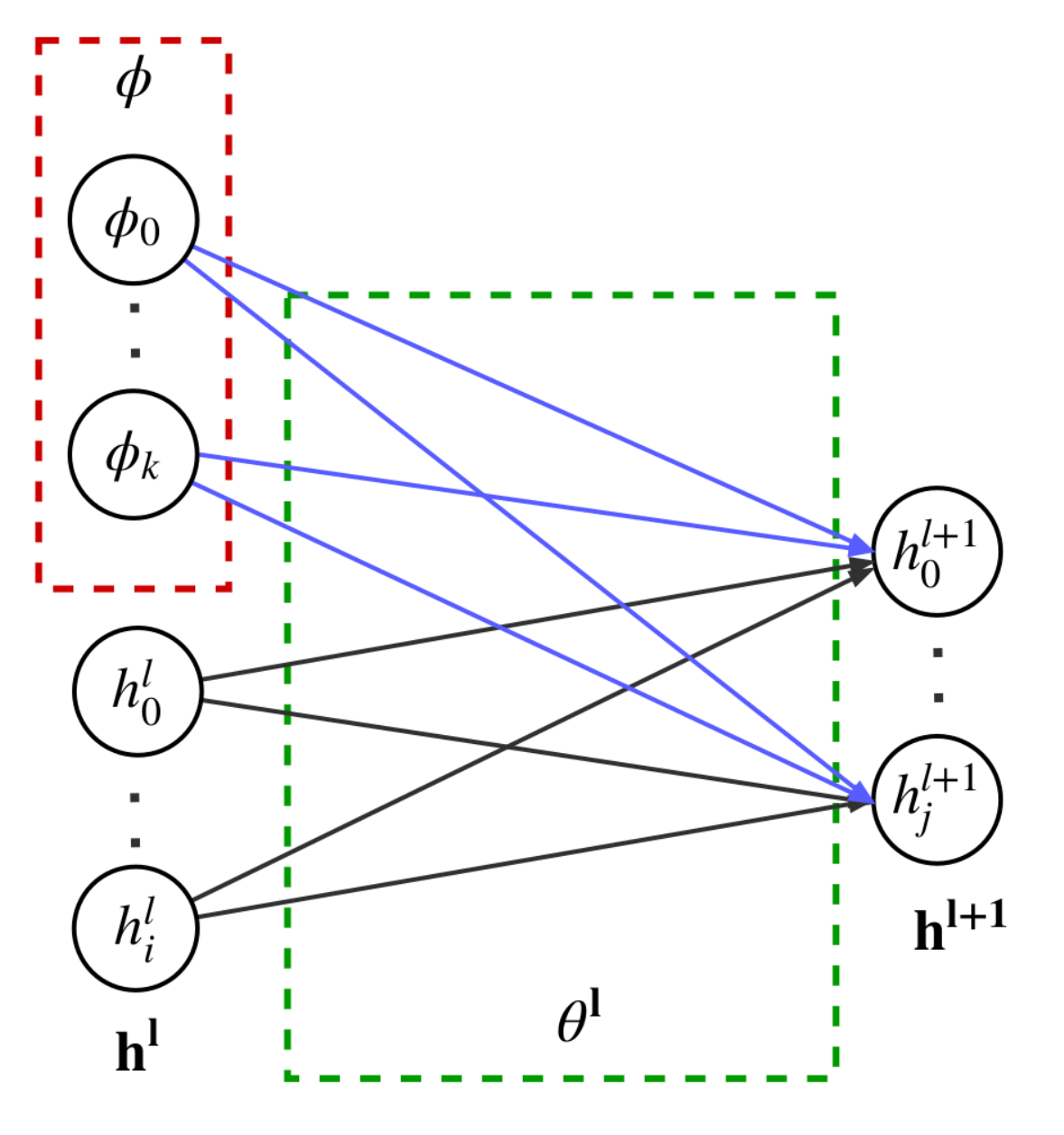}
		\caption{\textbf{Context adaptation.} A network layer $h^l$ is augmented with additional context parameters $\phi$ (red) initialised to $0$ before each adaptation step and updated by gradient descent during each inner loop and at test time. Network parameters $\theta$ (green) are only updated in the outer loop and shared across tasks. Hence, they stay fixed at test time. By initialising $\phi$ to 0, the network parameters associated with the context parameters (blue) do not affect the output of the layer before adaptation. After the first adaptation step they modulate the rest of the network to solve the new task.}
		\label{fig:overview}
\end{figure}

Since  $\phi$ is independent of the network input, we need to decide where and how to condition the network on them.
For an output node $h_i^{(l)}$ at a fully connected layer $l$, we can for example simply concatenate $\phi$ to the inputs of that layer:
\begin{equation} \label{eq:context_param_init}
h^{(l)}_i = g \left( \sum\limits_{j=1}^J 
\theta^{(l, h)}_{j,i} 
\ h^{(l-1)}_j + 
\sum\limits_{k=1}^K 
\theta^{(l, \phi)}_{k, i} \ 
\phi_{0,k} + b \right),
\end{equation}
where $g$ is a nonlinear activation function, $b$ is a bias parameter, $\theta^{(l, h)}_{j, i}$ are the weights associated with layer input $h^{(l-1)}_j$, and $\theta^{(l, \phi)}_{k, i}$ are the weights associated with the context parameter $\phi_{0,k}$.
This is illustrated in Figure \ref{fig:overview}.
In our experiments, for fully connected networks, we add the context parameter at the first layer, i.e., concatenate them to the input.
Other conditioning methods can be used with CAVIA as well:
e.g., for convolutional networks, we use {\it feature-wise linear modulation} FiLM \citep{perez2017film}, which performs an affine transformation on the feature maps.
Given context parameters $\phi$ and a convolutional layer that outputs $M$ feature maps $\{h_i\}_{i=1}^M$, FiLM linearly transforms each feature map $FiLM(h_i) = \gamma_i h_i + \beta $, where $\gamma, \beta \in \mathbb{R}^M$ are a function of the context parameters.
We use a fully connected layer $[\gamma, \beta] = \sum_{k=1}^K \theta^{(l, \phi)}_{k, i} \phi_{0, k} + b$ with the identity function at the output. 

\subsection{Context Parameter Initialisation} \label{sec:method_initialisation}

When learning a new task, $\phi$ have to be initialised to some value, $\phi_0$.
We argue that, instead of meta-learning this initialisation as well, a fixed $\phi_0$ is sufficient: in \eqref{eq:context_param_init}, if both $\theta^{(l, \phi)}_{j, i}$ and $\phi_0$ are meta-learned, the learned initialisation of $\phi$ can be subsumed into the bias parameter $b$, and $\phi_0$ can be set to a fixed value.
Hence, the initialisation of the context parameters does not have to be meta-learned and parameter copies are not required during training.
In our implementation we set the initial context parameter to a vector filled with zeros, $\phi_0=\mathbf{0}=[0, \dotsc, 0]^\top$.

Furthermore, not updating the context parameters $\phi$ in the outer loop allows for a more flexible and expressive gradient in the inner loop.
Consequently, CAVIA is more robust to the inner loop learning rate, $\alpha$ in \eqref{eq:inner_update}. 
Before an inner loop update, the part of the model associated with $\phi$ does not affect the output (since they are inputs and initialised at 0). 
During the inner update, only $\phi$ changes and can affect the output of the network at test time. 
Even if this update is large, the parameters $\theta^{(l, \phi)}_{k, i}$ that connect $\phi$ to the rest of the model (shown in blue in Figure \ref{fig:overview}), are automatically scaled during the outer loop. 
In other words, $\theta^{(l, \phi)}_{k, i}$  compensates in the outer loop for any excessively large inner loop update of $\phi$. 
However, doing large gradient updates in every outer loop update step as well would lead to divergence and numerical overflow. 
In Section \ref{sec:exp_regression}, we show empirically that the decoupling of learning $\phi$ and $\theta$ can indeed make CAVIA more robust to the initial learning rate compared to also learning the initialisation of the context parameters.

%% file: core/relatedwork.tex

\begin{table*}[t] \setlength{\tabcolsep}{4pt}
	\centering
	\begin{tabular}{| c || c | c c c c c c |} 
		\hline \multicolumn{1}{|c||}{} & \multicolumn{7}{|c|}{Number of Additional Input Parameters} \\
		Method & 0 & 1 & 2 & 3 & 4 & 5 & 50 \\ \hline \hline
		CAVIA  & - & $0.84 (\pm 0.06)$ & $0.21 (\pm 0.02)$ & $0.20 (\pm 0.02)$ & $\mathbf{0.19} (\pm 0.02)$ & $\mathbf{0.19} (\pm 0.02)$ & $\mathbf{0.19} (\pm 0.02)$\\
		MAML & $0.33 (\pm 0.02)$ & $0.29 (\pm 0.02)$  & $0.24 (\pm 0.02)$ & $0.24 (\pm 0.02)$  & $\mathbf{0.23} (\pm 0.02)$ &$\mathbf{0.23} (\pm 0.02)$ & $\mathbf{0.23} (\pm 0.02)$ \\ 
		\hline
	\end{tabular}
	\caption{Results for the sine curve regression task. Shown is the mean-squared error of CAVIA and MAML for varying number of input parameters, with $95\%$ confidence intervals in brackets.}
	\label{table:regression_results}
\end{table*}
\begin{figure*}[t]
	\centering
	\begin{subfigure}{0.3\textwidth}
		\includegraphics[width=\columnwidth]{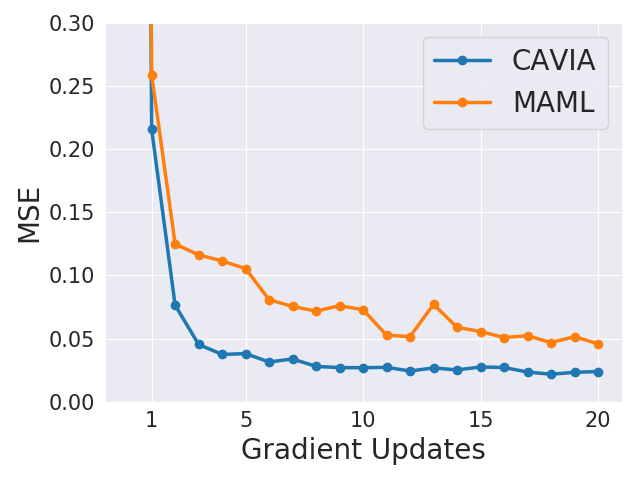}
		\caption{Test Performance}
		\label{fig:regression_update_steps}
	\end{subfigure}
	\begin{subfigure}{0.3\textwidth}
		\centering
		\includegraphics[width=\columnwidth]{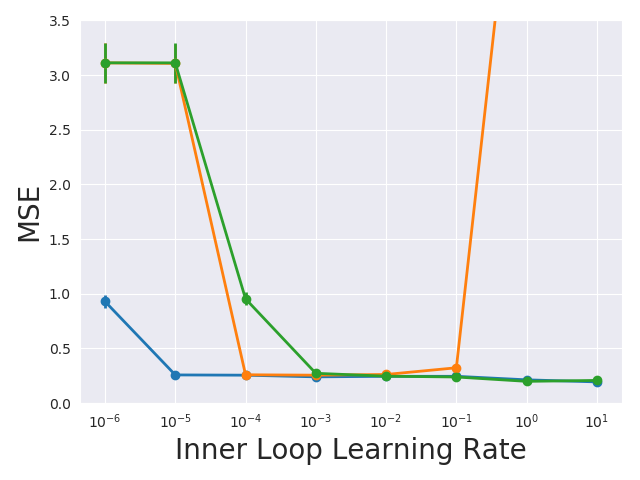}
		\caption{Learning Rates}
		\label{fig:learning_rates}
	\end{subfigure}
	\begin{subfigure}{0.3\textwidth}
		\centering
		\includegraphics[width=\columnwidth]{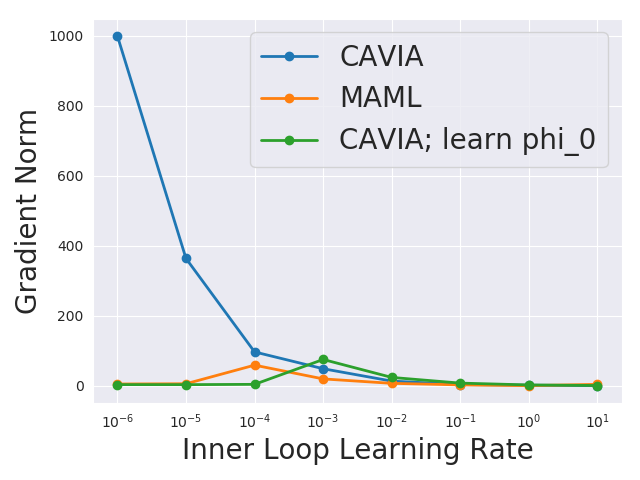}
		\caption{Gradient Norms}
		\label{fig:gradient_norms}
	\end{subfigure}
	\caption{Analysis of the sine curve experiments. (a) Test performance after several gradient steps (on the same batch) averaged over $1000$ unseen tasks. Both CAVIA and MAML continue to learn, but MAML performs worse and is less stable. (b) Test Performance after training with different inner loop learning rates. (c) CAVIA scales the model weights so that the inner learning rate is compensated by the context parameters gradients magnitude.}
	\label{fig:regression_results}
\end{figure*}

\section{Related Work}

One general approach to meta-learning is to learn the algorithm or update function itself \citep{schmidhuber1987evolutionary,bengio1992optimization,andrychowicz2016learning,ravi2016optimization}.
Another approach is gradient-based meta-learning, which learns a model initialisation such that at test time, a new task can be learned within a few gradient steps.
Examples are MAML \citep{finn2017model} and its probabilistic variants \citep{grant2018recasting,kim2018bayesian,finn2018probabilistic}; REPTILE \citep{nichol2018reptile}, which does not require second order gradient computation; and Meta-SGD \citep{li2017meta}, which learns the per-parameter inner loop learning rate.
The main difference to our work is that CAVIA adapts only a few parameters at test time, and these parameters determine only input context.

Closely related are MT-Nets \citep{lee2018gradient}, which learn {\it which} parameters to update in MAML.
MT-Nets learn: an M-Net which is a mask indicating which parameters to update in the inner loop, sampled (from a learned probability distribution) for each new task; and a T-net which learns a task-specific update direction and step size. 
CAVIA is a simpler, more interpretable alternative where the task-specific and shared parameters are disjoint sets.

Additional input biases to MAML were considered by \citet{finn2017one}, who show that this improves performance on a robotic manipulation setting. 
By contrast, we update {\it only} the context parameters in the inner loop, and initialise them to $0$ before adaptation to a new task. 
\citet{rei2015online} propose a similar approach in the context of neural language models, where a context vector represents the sentence that is currently being processed (see also the Appendix of \citet{finn2017model}).
Unlike CAVIA, this approach updates context parameters in the outer loop, i.e., it learns the initialisation of $\phi$.
This coupling of the gradient updates leads to a less flexible meta-update and is not as robust to the inner loop learning rate like CAVIA, as we show empirically in \ref{sec:exp_regression}.

\citet{silver2008inductive} proposed context features as a component of inductive transfer, using a predefined one-hot encoded task-specifying context as input to the network.
They show that this works better than learning a shared feature extractor and having separate heads for all tasks.
In this paper, we instead {\it learn} this contextual input from data of a new task.
Such context features can also be learned by a separate embedding network as in, e.g., \citet{oreshkin2018tadam} and \citet{garnelo2018conditional}, who use the task's training set to condition the prediction network.
CAVIA instead learns the context parameters via backpropagation through the same network used to solve the task.

Several methods learn to produce network weights from task-specific embeddings or labelled datapoints \citep{gordon2018meta, rusu2018meta}, which then operate on the task-specific inputs.
By contrast, we learn an embedding that modulates a fixed network, and is independent of the task-specific inputs during the forward pass.  
Specific to few-shot image classification, metric-based approaches learn to relate (embeddings of) labelled images and new instances of the same classes \citep{snell2017prototypical, sung2018learning}.
By contrast, CAVIA can be used for regression and reinforcement problems as well.
Other meta-learning methods are also motivated by the practical difficulties of learning in high-dimensional parameter spaces, and the relative ease of fast adaptation in lower dimensional space \citep[e.g.,][]{saemundsson2018meta,zhou2018deep}.

In the context of reinforcement learning, \citet{gupta2018meta} condition the policy on a latent random variable trained similarly to CAVIA, together with the reparametrisation trick (although they do not explicitly interpret these parameters as task embeddings).
This latent variable is sampled once per episode, and thus allows for structured exploration. 
Unlike CAVIA, they adapt the entire network at test time, which can be prone to overfitting.

%% file: core/experiments.tex

\section{Experiments} \label{sec:experiments}

In this section, we empirically evaluate CAVIA on regression, classification, and RL tasks. 
We show that: 
1) adapting a small number of input parameters (instead of the entire network) is sufficient to yield performance equivalent to or better than MAML,
2) CAVIA is robust to the task-specific learning rate and scales well without overfitting, and
3) an embedding of the task emerges in the context parameters solely via backpropagation.
Code is available at \url{https://github.com/lmzintgraf/cavia}.

\subsection{Regression} \label{sec:exp_regression}

\subsubsection{Sine Curves} 

We start with the regression problem of fitting sine curves from \citet{finn2017model}.
A task is defined by the amplitude and phase of the sine curve and generated by uniformly sampling the amplitude from $[0.1, 0.5]$ and the phase from $[0, \pi]$. 
For training, ten labelled datapoints (uniformly sampled from $x\in[-5, 5]$) are given for each task for the inner loop update, and we optimise a mean-squared error (MSE) loss.
We use a neural network with two hidden layers and $40$ nodes each.
The number of context parameters varies between $2$ and $50$.
Per meta-update we use a batch of $25$ tasks.
During testing we present the model with ten datapoints from $1000$ newly sampled tasks and measure MSE over 100 test points.

To allow a fair comparison, we add additional input biases to MAML (the same number as context parameters that CAVIA uses), an extension that was also done by \citet{finn2017one}.
These additional parameters are meta-learned together with the rest of the network.

\begin{figure}[t!]
	\centering
	\includegraphics[width=\columnwidth]{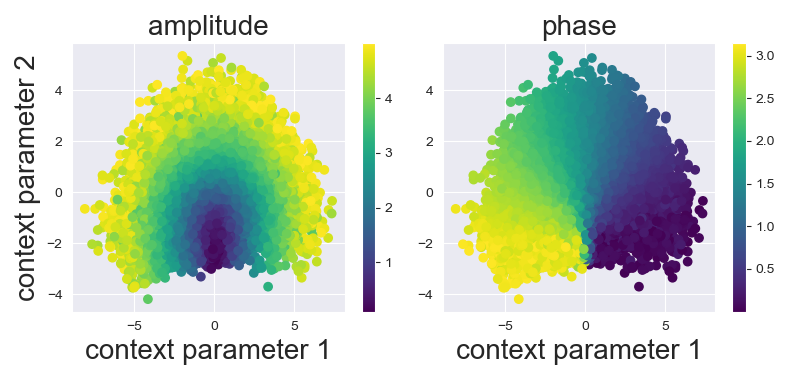}
	\caption{Visualisation of what the two context parameters learn on a new task. Shown is the value they take after $5$ gradient update steps on a new task. Each dot is one random task; its colour indicates the amplitude (left) or phase (right) of that task.}
	\label{fig:regression_correlations}
\end{figure}

Table 1 shows that CAVIA outperforms MAML even when MAML gets the same number of additional parameters, despite the fact that CAVIA adapts only $2$-$5$ parameters, instead of around $1600$. 
CAVIA's performance on the regression task correlates with how many variables are needed to encode the tasks. 
In these experiments, two parameters vary between tasks, which is exactly the context parameter dimensionality at which CAVIA starts to perform well (the optimal encoding is three dimensional, as phase is periodic).
This suggests CAVIA indeed learns task descriptions in the context parameters via backpropagation at test time.
Figure \ref{fig:regression_correlations} illustrates this by plotting the value of the learned inputs against the amplitude/phase of the task in the case of two context parameters. 
The model learns a smooth embedding in which interpolation between tasks is possible. 

We also test how well CAVIA can continue learning at test time, when more gradient steps are performed than during training. 
Figure \ref{fig:regression_update_steps} shows that CAVIA outperforms MAML even after taking several gradient update steps and is more stable, as indicated by the monotonic learning curve.

As described in Section \ref{sec:method_initialisation}, CAVIA can scale the gradients of the context parameters since they are inputs to the model and trained separately.
Figure~\ref{fig:learning_rates} shows the performance of CAVIA, MAML, and CAVIA when also learning the initialisation of $\phi$ (i.e., updating the context parameters in the outer loop), for a varying learning rate from $10^{-6}$ to $10$. 
CAVIA is robust to changes in learning rate while MAML performs well only in a small range. 
Figure~\ref{fig:gradient_norms} gives insight into how CAVIA does this: we plot the inner learning rate against the norm of the gradient of the context parameters at test time. 
The weights are adjusted so that lower learning rates bring about larger context parameter gradients and vice-versa.
MT-Nets \citep{lee2018gradient}, which learn which subset of parameters to adapt on a new task, are also robust to the inner-loop learning rate, but in a smaller range than CAVIA.\footnote{We do not show the numbers they report since we outperform them significantly, likely due to a different experimental protocol.}
Similarly, \citet{li2017meta} show that MAML can be improved by learning a parameter-specific learning rate, which, however, introduces a lot of additional parameters. 

\subsubsection{Image Completion}

\begin{figure}[t!]
	\centering
	\includegraphics[width=\columnwidth]{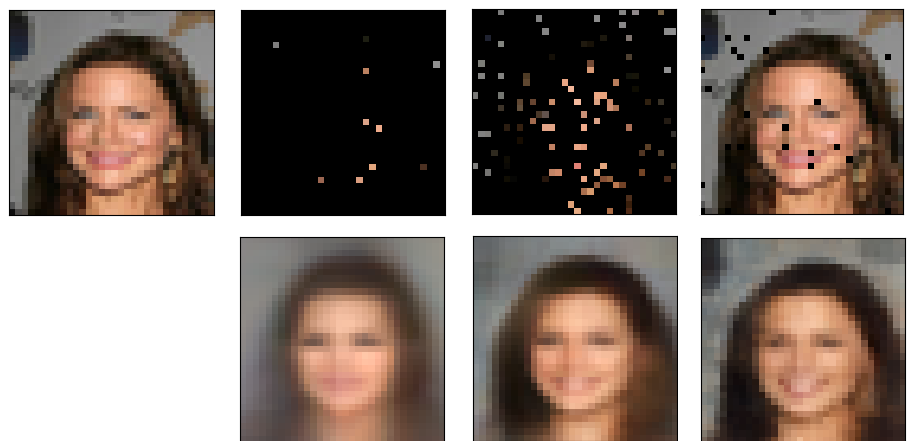}
	\caption{Image completion results on CelebA. Top row: true image on the left, and the training pixels for $10$, $100$, and $1000$ training points. Bottom row: prediction of CAVIA when $128$ context parameters were updated for $5$ gradient steps.}
	\label{fig:celeba_results}
\end{figure}
\begin{table}[t!] \setlength{\tabcolsep}{3.5pt}
	\centering
	\begin{tabular}{| c | c c c | c c c |} 
		\cline{2-7} \multicolumn{1}{c}{} & \multicolumn{3}{|c|}{Random Pixels} & \multicolumn{3}{|c|}{Ordered Pixels} \\
		\multicolumn{1}{c|}{}  & 10 & 100 & 1000 & 10 & 100 & 1000 \\ \hline \hline
		CNP$^*$  & 0.039 & 0.016  & 0.009 & 0.057 & $\mathbf{0.047}$ & 0.021 \\
		MAML & 0.040 & 0.017 & $\mathbf{0.006}$ & 0.055 & $\mathbf{0.047}$ &  0.007 \\ 
		CAVIA & $\mathbf{0.037}$ & $\mathbf{0.014}$ & $\mathbf{0.006}$ & $\mathbf{0.053}$ & $\mathbf{0.047}$ & $\mathbf{0.006}$  \\ 
		\hline
	\end{tabular}
	\caption{Pixel-wise MSE (for the entire image) for the image completion task on the CelebA data set. We test different number of available training points per image (10, 100, 1000). The trainig pixels are chosen either at random or ordered from the top-left corner to the bottom-right. (*Results from \citet{garnelo2018conditional})}
	\label{table:mse_celeba}
\end{table}

\begin{table*}[t!]
	\centering
	\begin{tabular}{| l | c | c |}
		\hline & \multicolumn{2}{|c|}{5-way accuracy} \\ 
		Method &  1-shot & 5-shot  \\ \hline \hline 
		Matching Nets \citep{vinyals2016matching} 	 & $46.6 \%$ & $60.0 \%$ \\
		Meta LSTM \citep{ravi2016optimization} 			& $43.44 \pm 0.77 \%$ & $60.60 \pm 0.71 \%$ \\
		Prototypical Networks \citep{snell2017prototypical} & $46.61 \pm 0.78 \%$ & $65.77 \pm 0.70\%$ \\
		\hline
		Meta-SGD \citep{li2017meta} 						& $50.47 \pm 1.87\%$ & $64.03 \pm 0.94\%$ \\
		REPTILE \citep{nichol2018reptile} 					& $49.97 \pm 0.32 \%$ & $65.99 \pm 0.58 \%$ \\
		MT-NET \citep{lee2018gradient} 					& $\mathbf{51.70} \pm 1.84 \%$ & - \\
		VERSA \citep{gordon2018meta} 			& $\mathbf{53.40} \pm 1.82\%$ & $\mathbf{67.37} \pm 0.86$ \\ 
		%
		%
		\hline
		\hline
		MAML (32) \citep{finn2017model} 					& $48.07 \pm 1.75 \%$ & $63.15 \pm 0.91 \%$       \\  
		MAML (64) 											 & $44.70 \pm 1.69\%$ &  $61.87 \pm 0.93\%$  \\  
		\hline
		CAVIA (32) 	 & $47.24 \pm 0.65\% $ & $59.05 \pm 0.54 \%$ \\  
		CAVIA (128)	& $49.84 \pm 0.68\% $ & $64.63 \pm 0.54\%$ \\  
		CAVIA (512)	& $\mathbf{51.82} \pm 0.65\%$ & $65.85 \pm 0.55\% $ \\  
		\hline
		CAVIA (512, first order)  & $49.92 \pm 0.68\%$ & $63.59 \pm 0.57\%$ \\  
		\hline
	\end{tabular}
	\caption{
		Few-shot classification results on the Mini-Imagenet test set (average accuracy with $95\%$ confidence intervals on a random set of $1000$ tasks).
		For MAML, we show the results reported by \citet{finn2017model}, and when using a larger network (results obtained with the author's open sourced code and unchanged hyperparameters except the number of filters). 
		These results show that CAVIA is able to scale to larger networks without overfitting, and outperforms MAML by doing so. 
		We also include other CNN-based methods with similar experimental protocol. 
		Note however that we did not tune CAVIA to compete with these methods, but focus on the comparison to MAML in this experiment.
	}
	\label{table:classification_results} \vspace{-0.5cm}
\end{table*}

To evaluate CAVIA on a more challenging regression task, we consider image completion \citep{garnelo2018conditional}. 
The task is to predict pixel values from coordinates, i.e., learn a function $f: [0,1]^2 \rightarrow [0, 1]^3$ (for RGB values) which maps $2$D pixel coordinates $x\in [0,1]^2$ to pixel intensities $y \in [0,1]^3$. 
An individual picture is considered a single task, and we are given a few pixels as a training set $\mathcal{D}^\text{train}$ and use the entire image as the test set $\mathcal{D}^\text{test}$ (including the training set). 
We train CAVIA on the CelebA \citep{liu2015faceattributes} training set, perform model selection on the validation set, and evaluate on the test set.

\citet{garnelo2018conditional} use an MLP encoder with three hidden layers and $128$ nodes each, a $128$-dimensional embedding size, and a decoder with five hidden layers with $128$ nodes each.
To allow a fair comparison, we therefore choose a context vector of size $128$, and use an MLP with five hidden layers ($128$ nodes each) for the main network.
We chose an inner-learning rate of $1.0$ without tuning.
To train MAML, we use the same five-layer MLP network including $128$ additional input biases, and an inner-loop learning rate of $0.1$ (other tested learning rates: $1.0$, $0.01$).
Both CAVIA and MAML were trained with five inner-loop gradient updates.

Table \ref{table:mse_celeba} shows the results in terms of pixel-wise MSE for different numbers of training pixels ($k=10,100,1000$ shot), and for the case of randomly selected pixels and ordered pixels (i.e., selecting pixels starting from the top left of the image). 
CAVIA outperforms CNPs and MAML in most settings. 
Figure \ref{fig:celeba_results} shows an example image reconstruction produced by CAVIA (see Appendix \ref{appendix:celeba} for more results).

These results show that it is possible to learn an embedding only via backpropagation and with far fewer parameters than when using a separate embedding network.

\subsection{Classification} \label{sec:experiments_classificaton}

\begin{figure*}[t!]
	\centering
	\begin{subfigure}{\columnwidth}
		\centering
		\includegraphics[width=0.65\columnwidth]{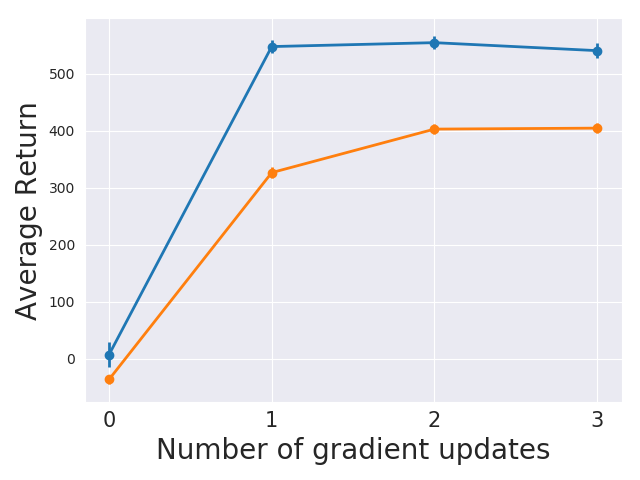}
		\caption{Direction}
		\label{fig:cheetah_dir}
	\end{subfigure}
	\vspace{-0.2cm}
	\begin{subfigure}{\columnwidth}
		\centering
		\includegraphics[width=0.6\columnwidth]{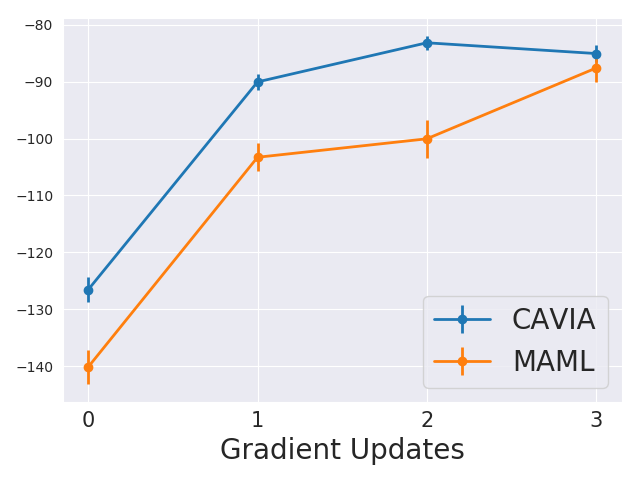}
		\caption{Velocity}
		\label{fig:cheetah_vel}
	\end{subfigure}
	\caption{Performance of CAVIA and MAML on the RL Cheetah experiments. Both agents were trained to perform one gradient update, but are evaluated for several update steps. Results are averaged over $40$ randomly selected tasks.}
	\label{fig:cheetah}
\end{figure*}

To evaluate how well CAVIA can scale to problems that require larger networks, we test it on the few-shot image classification benchmark Mini-Imagenet \citep{ravi2016optimization}. 
In $N$-way $K$-shot classification, a task is a random selection of $N$ classes, for each of which the model gets to see $K$ examples. 
From these it must learn to classify unseen images from the $N$ classes.
The Mini-Imagenet dataset consists of 64 training classes, 12 validation classes, and 24 test classes.
During training, we generate a task by selecting $N$ classes at random from the $64$ classes and training the model on $K$ examples of each, i.e., a batch of $N\times K$ images.
The meta-update is done on a set of unseen images of the same classes.

On this benchmark, MAML uses a network with four convolutional layers with $32$ filters each and one fully connected layer at the output \citep{finn2017model}.
We use the same network architecture, but between $32$ and $512$ filters per layer.
We use $100$ context parameters and add a  FiLM layer  that conditions on these after the third convolutional layer and whose parameters are meta-learned with the rest of the network, i.e., they are part of $\theta$.
All our models were trained with two gradient steps in the inner loop and evaluated with two gradient steps.
Following \cite{finn2017model}, we ran each experiment for $60,000$ meta-iterations and selected the model with the highest validation accuracy for evaluation on the test set.

Table \ref{table:classification_results} shows our results on Mini-Imagenet held-out test data for $5$-way $1$-shot and $5$-shot classification. 
Our smallest model ($32$ filters) underperforms MAML (within the confidence intervals), and our largest model ($512$ filters) clearly outperforms MAML.
We also include results for the first order approximation of our largest models, where the gradient with respect to $\theta$ is not backpropagated through the inner loop update of the context parameters $\phi$.
As expected, this results in a lower accuracy (a drop of $2\%$) , but we still outperform MAML.

CAVIA benefits from increasing model expressiveness:
since we only adapt the context parameters in the inner loop per task, we can substantially increase the network size without overfitting during the inner loop update.
We tested scaling up MAML to a larger network size as well (see Table \ref{table:classification_results}), but found that this hurt accuracy, which was also observed by \citet{mishra2018simple}.

Note that we focus on the comparison to MAML in these experiments, since our goal is to show that CAVIA can scale to larger networks without overfitting compared to MAML, and can be used out-of-the-box for classification problems as well.
We did not tune CAVIA in terms of network architecture or other hyperparameters, but only varied the number of filters at each convolutional layer.
 The best performing method with similar architecture and experimental protocol is VERSA \citep{gordon2018meta}, which learns to produce weights of the classifier, instead of modulating the network. 
The current state-of-the-art results in Mini-Imagenet is (to the best of our knowledge) the method LEO \citet{rusu2018meta} who use pre-trained feature representations from a deep residual network \citep{he2016deep} and a different experimental protocol.
CAVIA can be used with such embeddings as well, and we expect an increase in performance.

In conclusion, CAVIA can achieve much higher accuracies than MAML by increasing the network size, without overfitting.
Our results are obtained by adjusting only $100$ parameters at test time (instead of $>30,0000$ like MAML), which embed the five different classes of the current task.

\subsection{Reinforcement Learning} \label{sec:exp_rl}

To demonstrate the versatility of CAVIA, we also apply it to two high dimensional reinforcement learning MuJoCo \citep{todorov2012mujoco} tasks using the setup of \citet{finn2017model}.
In the first experiment, a Cheetah robot must run in a particular, randomly chosen direction (forward/backward), and receives as reward its speed in that direction.
In the second experiment, the Cheetah robot must run at a particular velocity, chosen uniformly at random between $0.0$ and $2.0$.
The agent's reward is the negative absolute value between its current and the target velocity.
Each rollout has a length of $200$, and we use $20$ rollouts per gradient step during training, and a meta-batchsize of $40$ tasks per outer update.
As in \citet{finn2017model}, our agents are trained for one gradient update, using policy gradient with generalised advantage estimation \citep{schulman2015high} in the inner loop and TRPO \citep{schulman2015trust} in the outer loop update.
Following the protocol of \citet{finn2017model}, both CAVIA and MAML were trained for up to 500 meta-iterations, and the models with the best average return during training were used for evaluation.
For these tasks, we use $50$ context parameters for CAVIA and an inner-loop learning rate of $10$.
We found that starting with a higher learning rate helps for RL problems, since the policy update in the outer loop has a stronger signal from the context parameters.

Figure \ref{fig:cheetah} shows the performance of the CAVIA and MAML agents at test time, after up to three gradient steps (averaged over $40$ randomly selected test tasks).
Both models keep learning for several updates, although they were only trained for one update step. 
CAVIA outperforms MAML on both domains after one gradient update step, while updating only $50$ parameters at test time per task compared to $>10,000$.
For the Cheetah Velocity experiment, MAML catches up after three gradient update steps.

%% file: core/discussion.tex
\section{Conclusion and Future Work}

CAVIA is a meta-learning approach that separates the model into task-specific context parameters and parameters that are shared across tasks.
We demonstrated experimentally that CAVIA is robust to the inner loop learning rate and yields task embeddings in the context parameters.
CAVIA outperforms MAML on challenging regression, classification, and reinforcement learning problems, while adapting fewer parameters at test time and being less prone to overfitting.

We are interested in extending our experimental evaluation to settings with multi-modal task distributions, as well as settings where more generalisation beyond task identification is necessary at test time.
One possible approach here is to combine CAVIA with MAML-style updates in the future: i.e., having two separate inner loops (producing $\phi_i$ and $\theta_i$), and one outer loop.

We are interested in extending CAVIA to more challenging RL problems and exploring its role in allowing for smart exploration in order to identify the task at hand, for example building on the work of \citet{gupta2018meta} who use probabilistic context variables, or \citet{stadie2018some}, who propose E-MAML for RL problems, an extension for MAML which accounts for the effect of the initial sampling distribution (policy) before the inner-loop update.

%% file: core/appendix.tex
\appendix

\begin{center}
{\textbf{Fast Context Adaptation via Meta-Learning}} \\ \vspace{0.3cm}
{\centering \Large Supplementary Material} 
\end{center}
	
\section{Pseudocode} \label{eq:appendix_code}

\begin{algorithm}[H] \label{alg:MAML}
	\caption{CAVIA for Supervised Learning}
	\begin{algorithmic}[1] 
		\REQUIRE Distribution over tasks $p(\mathcal{T})$
		\REQUIRE Step sizes $\alpha$ and $\beta$
		\REQUIRE Initial model $f_{\phi_0, \theta}$ with $\theta$ initialised randomly and $\phi_0=0$ 
		\WHILE{not done}
		\STATE Sample batch of tasks $\mathbf{T} = \{\mathcal{T}_i\}_{i=1}^N$ where $\mathcal{T}_i\sim p$
		\FORALL{$\mathcal{T}_i\in\mathbf{T}$}
		\STATE $\mathcal{D}_i^\text{train}, \mathcal{D}_i^\text{test} \sim q_{\mathcal{T}_i}$
		\STATE $\phi_0 = 0$ 
		\STATE $\phi_i = 
		\phi_0 - 
		\alpha
		\nabla_{\phi}
		\frac{1}{M_i^\text{train}}
		\sum\limits_{(x, y) \in \mathcal{D}_i^\text{train}} 
		\mathcal{L}_{\mathcal{T}_i} (f_{\phi_0, \theta}(x), y)$
		\ENDFOR
		\STATE $	\theta \leftarrow \theta - 
		\beta
		\nabla_\theta 
		\frac{1}{N}
		\sum\limits_{\mathcal{T}_i \in \mathbf{T}}
		\frac{1}{M_i^\text{test}}
		\sum\limits_{(x, y) \in \mathcal{D}_i^\text{test}}
		\mathcal{L}_{\mathcal{T}_i} (f_{\phi_i, \theta}(x, y))$ 
		\ENDWHILE
	\end{algorithmic}
\end{algorithm}

\begin{algorithm}[H] \label{alg:CAVIA}
	\caption{CAVIA for RL}
	\begin{algorithmic}[1]
		\REQUIRE Distribution over tasks $p(\mathcal{T})$
		\REQUIRE Step sizes $\alpha$ and $\beta$
		\REQUIRE Initial policy $\pi_{\phi_0, \theta}$ with $\theta$ initialised randomly and $\phi_0=0$
		\WHILE{not done}
		\STATE Sample batch of tasks $\mathbf{T} = \{\mathcal{T}_i\}_{i=1}^N$ where $\mathcal{T}_i\sim p$
		\FORALL{$\mathcal{T}_i\in\mathbf{T}$}
		\STATE Collect rollout $\tau_i^\text{train}$ using $\pi_{\phi_0, \theta}$
		\STATE $\phi_i = 
		\phi_0 + 
		\alpha
		\nabla_\phi 
		\tilde{\mathcal{J}}_{\mathcal{T}_i} (\tau_i^\text{train}, \pi_{\phi_0, \theta})$ 
		\STATE Collect rollout $\tau_i^\text{test}$ using $\pi_{\phi_i, \theta}$
		\ENDFOR
		\STATE $	\theta \leftarrow \theta + 
		\beta
		\nabla_\theta 
		\frac{1}{N}
		\sum\limits_{\mathcal{T}_i \in \mathbf{T}}
		\tilde{\mathcal{J}}_{\mathcal{T}_i} (\tau_i^\text{test}, \pi_{\phi_i, \theta})$ 
		\ENDWHILE
	\end{algorithmic}
\end{algorithm}

\section{Practical Tips}

\subsection{Implementation}

The context parameters $\phi$ can be added to any network, and do not require direct access to the rest of the network weights like MAML. In PyTorch this can be done as follows.
To add CAVIA parameters to a network, it is necessary to first initialise them to zero when the model is initialised:
{\small \begin{spverbatim}
		self.context_params = torch.zeros(size=[self.num_context_params],
		requires_grad=True)
\end{spverbatim}}

Add a way to reset the context parameters to zero (e.g., a method that just does the above).
During the forward pass, add the context parameters to the input by concatenating it (when using a fully connected network):
{\small \begin{spverbatim}
		x = torch.cat((x, 
		self.context_params.expand(x.shape[0], -1)), dim=1)
\end{spverbatim}}
(This is for fully connected networks. We refer the reader to our implementation for how to use FiLM to condition CNNs.)
To correctly set the computation graph for the outer loop, it is necessary to assign the context parameters manually with their gradient. In the inner loop, compute the gradient:
{\small \begin{spverbatim}
		grad = torch.autograd.grad(task_loss, model.context_params, 
		create_graph=True)[0]
\end{spverbatim}}
The option {\it create\_graph} will make sure that you can take the gradient of {\it grad} again.
Then, update the context parameters using one gradient descent step
{\small \begin{spverbatim}
		model.context_params = model.context_params - lr_inner * grad
\end{spverbatim}}
If you now do another forward pass and compute the gradient of the model parameters $\theta$ (for the outer loop), these will include higher order gradients because {\it grad} above includes gradients of $\theta$, and because we kept the computation graph via the option {\it grad}.
To see how to train CAVIA and aggregate the meta-gradient over several tasks, see our implementation at [blinded; see supplementary material].

\subsection{Hyperparameter Selection}

The choice of network architecture/size and context parameters can be guided by domain knowledge. E.g., for the few-shot image classification problem, an appropriate model is a deep convolutional model. For the context parameters, it is important to make sure they are not underparameterised. CAVIA can deal with larger than necessary context parameters (see Table \ref{table:regression_results}), although it might start overfitting in the inner loop at some point (we have not experiences this in practise).
Regarding learning rates, we always started with an inner loop learning rate of $1$ and the Adam optimiser with the standard learning rate of $0.001$ for the outer loop.

For CNNs, we found that adding the context parameters not at the input layer, but after several (in our case after the third out of four) convolutions works best. We believe this is because the lower-level features that the first convolutions extract are useful for any image classification task, and we only want our task embedding to influence the activations at the deeper layers. In our experiments we used a FiLM network with no hidden layers. We tried deeper versions, but this resulted in inferior performance. 

We also tested to add context parameters at several layers instead of only one. However, in our experience this resulted in similar (regression and RL) or worse (in the case of CNNs) performance. 

\begin{figure*}[h!]
	\centering
	\begin{subfigure}{0.3\textwidth}
		\includegraphics[width=\textwidth]{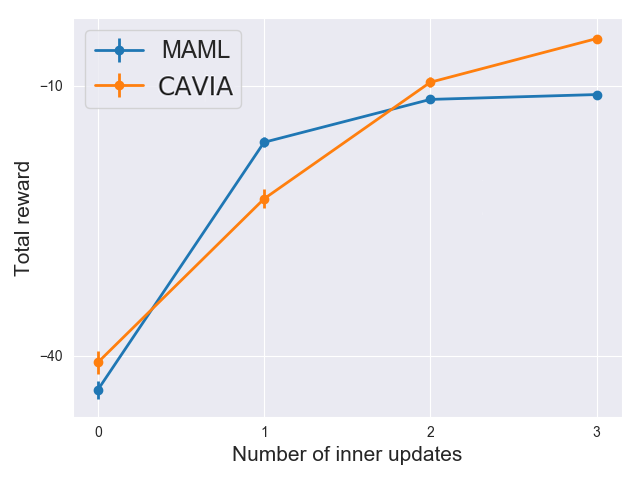}
		\caption{Multiple gradient updates.}
		\label{fig:rl_multiple_updates}
	\end{subfigure}
	\hspace{0.1cm}
	\begin{subfigure}{0.3\textwidth}
		\centering
		\includegraphics[width=\textwidth]{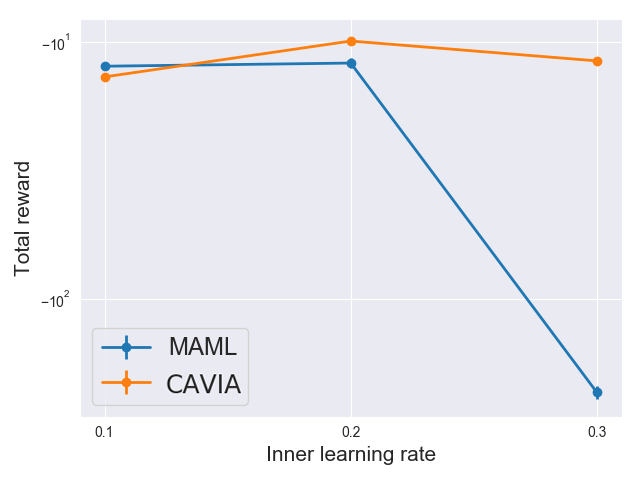}
		\caption{Learning rate comparison.}
		\label{fig:rl_learning_rate_comparison}
	\end{subfigure}
	\begin{subfigure}{0.3\textwidth}
		\centering
		\includegraphics[width=\textwidth]{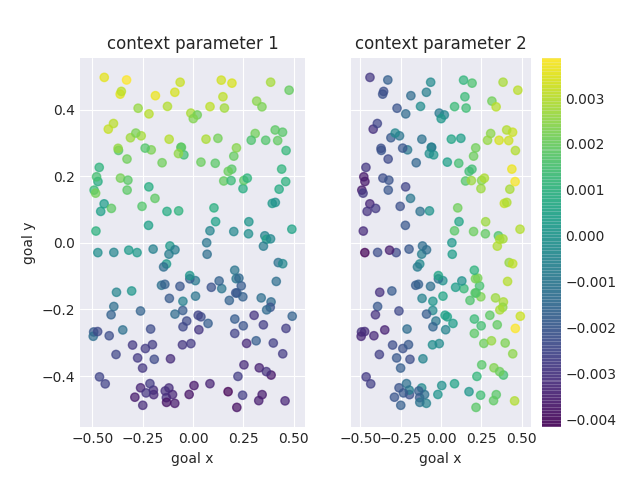}
		\caption{Goal position embedding.}
		\label{fig:navigation_embedding}
	\end{subfigure}
	\caption{Results for the $2$D navigation reinforcement learning problem.}
\end{figure*}

\section{Experiments}

\subsection{Classification: Details} \label{appendix:exp_classification}

For Mini-Imagenet, our model takes as input images of size $84\times84\times3$ and has $5$ outputs, one for each class.
The model has four modules that each consist of: 
a $2D$ convolution with a $3\times3$ kernel, 
padding $1$ and $128$ filters, 
a batch normalisation layer, 
a max-pooling operation with kernel size $2$, 
if applicable a FiLM transformation (only at the third convolution, details below), 
and a ReLU activation function. 
The output size of these four blocks is $5\times5\times128$, which we flatten to a vector and feed into one fully connected layer.
The FiLM layer itself is a fully connected layer with inputs $\phi$ and a $256$-dimensional output and the identity function at the output.
The output is divided into $\gamma$ and $\beta$, each of dimension $128$, which are used to transform the filters that the convolutional operation outputs.
The context vector is of size $100$ (other sizes tested: $50$, $200$) and is added after the third convolution (other versions tested: at the first, second or fourth convolution).

The network weights are initialised using \citet{he2015delving}, the bias parameters are initialised to zero (except at the FiLM layer).
We use the Adam optimiser for the meta-update step with an initial learning rate of $0.001$. 
This learning rate is annealed every $5,000$ steps by multiplying it by $0.9$.
The inner learning rate is set to $0.1$ (others tested: $1.0$, $0.01$).
We use a meta batchsize of $4$ and $2$ tasks for $1$-shot and $5$-shot classification respectively.
For the batch norm statistics, we always use the current batch -- also during testing. I.e., for $5$-way $1$-shot classification the batch size at test time is $5$, and we use this batch for normalisation.

\subsection{Reinforcement Learning: Additional Experiments} \label{appendix:exp_rl}

We also perform reinforcement learning experiments on the simple 2D Navigation task of \citet{finn2017model}. 
The agent moves in a 2D world using continuous actions and at each timestep is given a negative reward proportional to its distance from a pre-defined goal position. Each task has a new unknown goal position.  

We follow the same procedure as \citet{finn2017model}. 
Goals are sampled from an interval of $(x,y)=[-0.5,0.5]$. 
At each step we sample 20 tasks for both the inner and outer loops and testing is performed on 40 new unseen tasks. 
We learn for 500 iterations and optimise for one gradient update in the inner loop. 
The best performing policy during training is then presented with new test tasks and allowed two gradient updates. 
For each update, the total reward over 20 rollouts per task is measured.
We use a two-layer network with 100 units per layer and ReLU nonlinearities to represent the policy and a linear value function approximator. 
For CAVIA we use five context parameters at the input layer.

Figure \ref{fig:rl_multiple_updates} shows that the two methods are highly competitive. 
We think that the similar performance is mostly due to a ceiling effect, since the domain is relatively simple.
Notably, CAVIA adapts only five parameters at test time, whereas MAML adapts around $10,000$. 
Figure \ref{fig:rl_learning_rate_comparison}, which plots performance for several learning rates (at test time, after two gradient updates), shows that CAVIA is again less sensitive to the inner loop learning rate. 
Only when using a learning rate of $0.1$ is MAML competitive in performance.\footnote{For MAML we halve the learning rate after the first gradient update, following \citet{finn2017model}.}

As with regression, the optimal task embedding is low dimensional enough to plot. We therefore apply CAVIA with two context parameters and plot how these correlate with the actual position of the goal for 200 test tasks.
Figure \ref{fig:navigation_embedding} shows that the context parameters obtained after two policy gradient updates represent a disentangled embedding of the actual task. Specifically, context parameter 1 encodes the $y$ position of the goal, while context parameter 2 encodes the $x$ position. Hence, CAVIA can learn compact interpretable task embeddings via backpropagation through the inner loss. 

\subsection{Additional CelebA Image Completion Results} \label{appendix:celeba}

The following images show additional results for the CelebA image completion task.

\begin{figure*}[t]
	\centering
	\includegraphics[width=\textwidth]{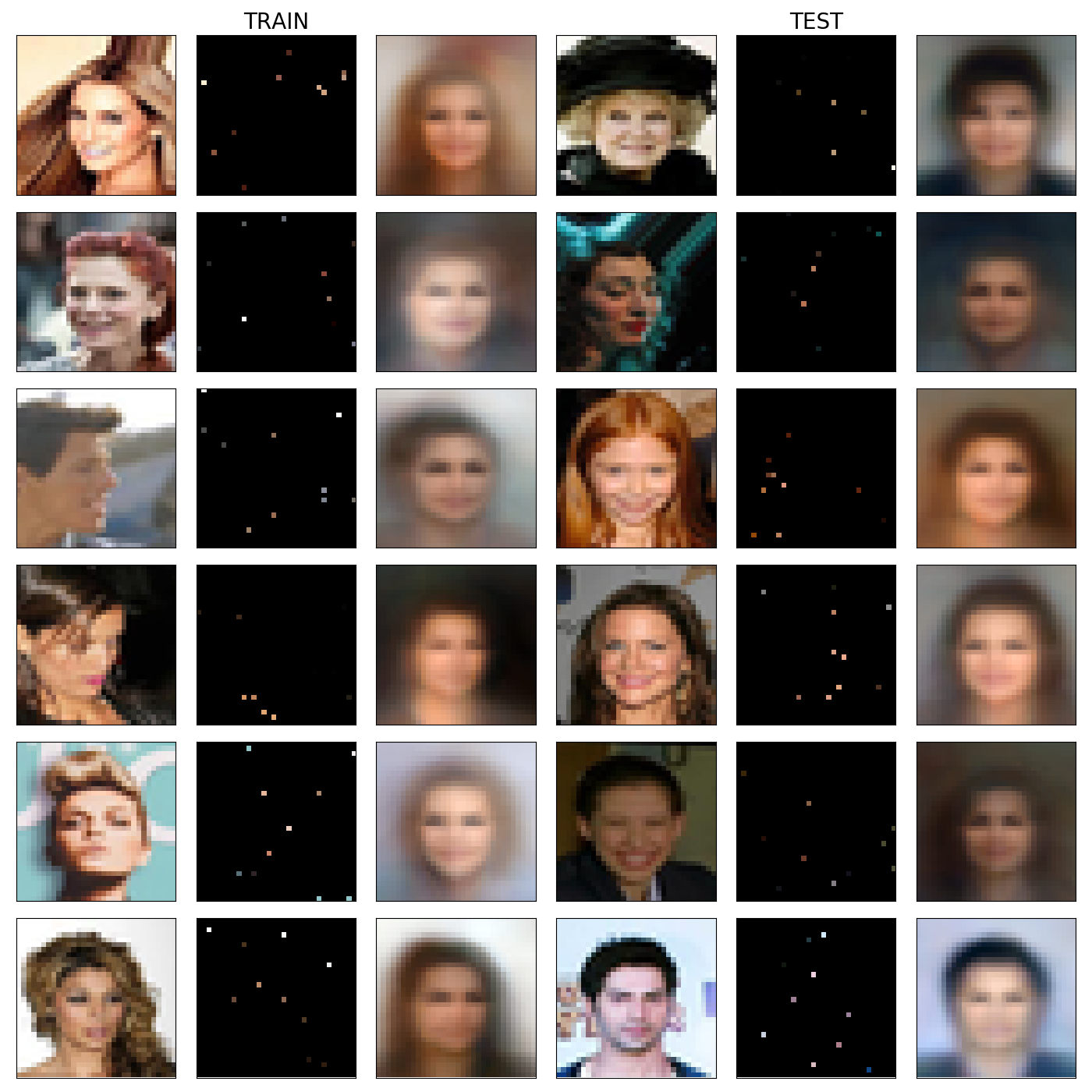}
	\caption{Additional image completion results for the CelebA image completion problem, when $k=10$ pixels are given.}
\end{figure*}

\begin{figure*}[t]
	\centering
	\includegraphics[width=\textwidth]{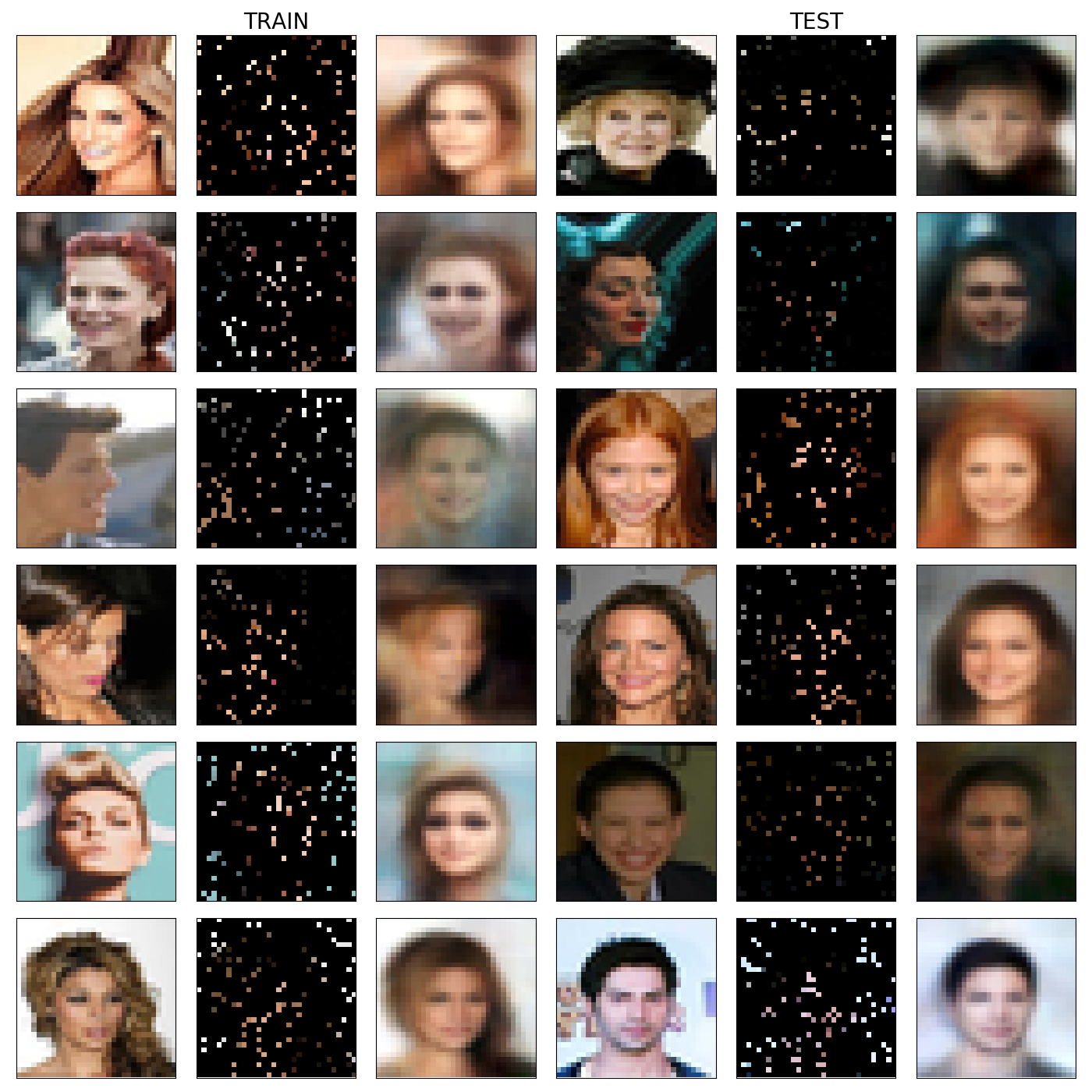}
	\caption{Additional image completion results for the CelebA image completion problem, when $k=10$ pixels are given.}
\end{figure*}

\begin{figure*}[t]
	\centering
	\includegraphics[width=\textwidth]{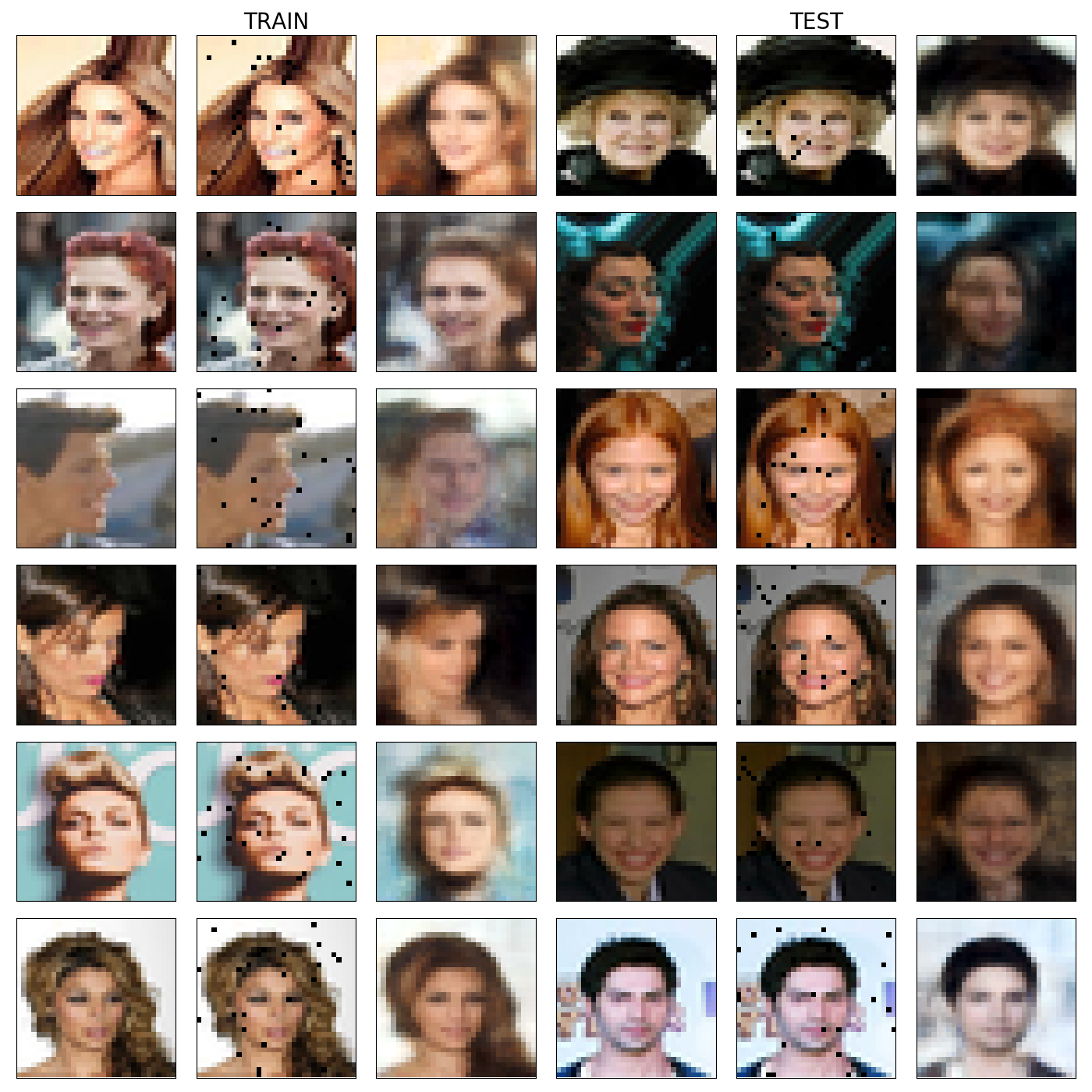}
	\caption{Additional image completion results for the CelebA image completion problem, when $k=10$ pixels are given.}
\end{figure*}